\documentclass[10pt,twocolumn,letterpaper]{article}

\usepackage{iccv}
\usepackage{times}
\usepackage{epsfig}
\usepackage{graphicx}
\usepackage{amsmath}
\usepackage{amssymb}

\usepackage{floatrow}
\usepackage{caption}
\usepackage{subcaption}
\usepackage{multicol}
\usepackage{multirow}
\usepackage{makecell}
\usepackage{graphicx}
\usepackage{xcolor}
\usepackage{booktabs}
\usepackage{amsmath}
\usepackage{amssymb}
\usepackage{wrapfig}
\usepackage{tabulary}
\usepackage{fontawesome}

\newcommand{\x}{{\times}} 
\newcommand{\deh}[1]{\textcolor{gray}{#1}}
\newcolumntype{x}[1]{>{\centering\arraybackslash}p{#1.5pt}}

\newlength\savewidth\newcommand\shline{\noalign{\global\savewidth\arrayrulewidth\global\arrayrulewidth1pt}\hline\noalign{\global\arrayrulewidth\savewidth}}
\newcommand{\tablestyle}[2]{\setlength{\tabcolsep}{#1}\renewcommand{\arraystretch}{#2}\centering\small}

\usepackage[pagebackref=true,breaklinks=true,letterpaper=true,colorlinks,bookmarks=false]{hyperref}
\definecolor{citecolor}{HTML}{0071BC}
\definecolor{linkcolor}{HTML}{ED1C24}
\usepackage[british, english, american]{babel}

\usepackage[capitalize]{cleveref}
\crefname{section}{Sec.}{Secs.}
\Crefname{section}{Section}{Sections}
\Crefname{table}{Table}{Tables}
\crefname{table}{Tab.}{Tabs.}

\iccvfinalcopy 

\def\iccvPaperID{5475} 

\ificcvfinal\fi

\begin{document}

\title{Pixel-Wise Contrastive Distillation}

\author{Junqiang Huang\textsuperscript{$\dagger$ $\ddagger$}  \quad Zichao Guo\textsuperscript{$\dagger$}\\
Shopee\\
{\tt\small jq.huang.work@gmail.com}}

\maketitle

\ificcvfinal\thispagestyle{empty}\fi

\renewcommand{\thefootnote}{\fnsymbol{footnote}}
\footnotetext[2]{Equal contribution}
\footnotetext[3]{Corresponding author}
\renewcommand{\thefootnote}{}

\renewcommand*{\thefootnote}{\arabic{footnote}}

\begin{abstract}
   We present a simple but effective pixel-level self-supervised distillation framework friendly to dense prediction tasks. Our method, called Pixel-Wise Contrastive Distillation (PCD), distills knowledge by attracting the corresponding pixels from student's and teacher's output feature maps. PCD includes a novel design called SpatialAdaptor which ``reshapes'' a part of the teacher network while preserving the distribution of its output features. Our ablation experiments suggest that this reshaping behavior enables more informative pixel-to-pixel distillation. Moreover, we utilize a plug-in multi-head self-attention module that explicitly relates the pixels of student's feature maps to enhance the effective receptive field, leading to a more competitive student. PCD outperforms previous self-supervised distillation methods on various dense prediction tasks. A backbone of \mbox{ResNet-18-FPN} distilled by PCD achieves $37.4$ AP$^\text{bbox}$ and $34.0$ AP$^\text{mask}$ on COCO dataset using the detector of \mbox{Mask R-CNN}. We hope our study will inspire future research on how to pre-train a small model friendly to dense prediction tasks in a self-supervised fashion. Our implementation is availvable at \url{https://github.com/allo-rene/pcd}.
   
   
\end{abstract}

\section{Introduction}

Self-supervised learning (SSL) has emerged as a promising pre-training method due to its remarkable progress on various computer vision tasks \cite{moco,simclr,byol,swav,mae,simmim}. Models pre-trained by SSL methods attain transfer performance akin to, or even surpassing that of their supervised pre-trained counterparts. However, this advancement of SSL appears to be confined to larger models. Small models, such as \mbox{ResNet-18} \cite{resnet}, exhibit inferior linear probing accuracy as reported in \cite{seed,bingo,distill_on_the_fly}. Considering the necessity of small models for edge devices or resource constraint regime, it is much essential to tackle this problem. 

Recently, the performance lag of small models has been effectively alleviated by \textit{self-supervised distillation} \cite{cluster_fit,boosting_ssl_via_kt,compress,seed,disco,bingo,distill_on_the_go,distill_on_the_fly,rekd,tinyvit}, where teachers' (large pre-trained models) knowledge is transferred \cite{model_compression,access_unlabeled_data,need_to_be_deep,hinton_kd} to students (small models) in a self-supervised learning fashion. Self-supervised distillation methods yield competitive performance for small models, especially on classification tasks (\eg, fine-grained and few-shot classification). Nevertheless, their improvement on dense prediction tasks like object detection and semantic segmentation is less pronounced than on classification tasks. This imbalance seemingly suggests that the favorable representations learned by teachers can only be \textit{partially} transferred to students. A natural question arises: what obstructs students from inheriting the knowledge advantageous to dense prediction tasks? In this study, we seek the answers to this question from the following aspects.


First, the distillation signals of current self-supervised distillation methods are mostly at image-level, while the rich \textit{pixel-level knowledge} is yet to be utilized. We argue that it is inefficient for small models to learn representations good for dense prediction tasks from image-level supervision\footnote{On the other hand, large models pre-trained by image-level SSL methods like \mbox{MoCo} can be quite competitive on dense prediction tasks.}. Driven by this, we here present a simple but effective pixel-level self-supervised distillation framework, \textbf{Pixel-Wise Contrastive Distillation} (PCD), which extends the idea of contrastive learning \cite{contrastive_learning} by incorporating pixel-level knowledge distillation. PCD attracts the \textit{corresponding} pixels from the student's and the teacher's output feature maps and separates the student's pixels and the negative pixels of a memory queue \cite{moco}. With pixel-level distillation signal, PCD enables more efficient and adequate transfer of knowledge from the teacher to the student. 



\begin{figure*}

    \ffigbox[\FBwidth]{%
        \begin{subfloatrow}
        \ffigbox[\FBwidth]{\caption{Architecture of Pixel-Wise Contrastive Distillation\label{fig:pcd_architecture}}}     {\includegraphics[width=.47\textwidth]{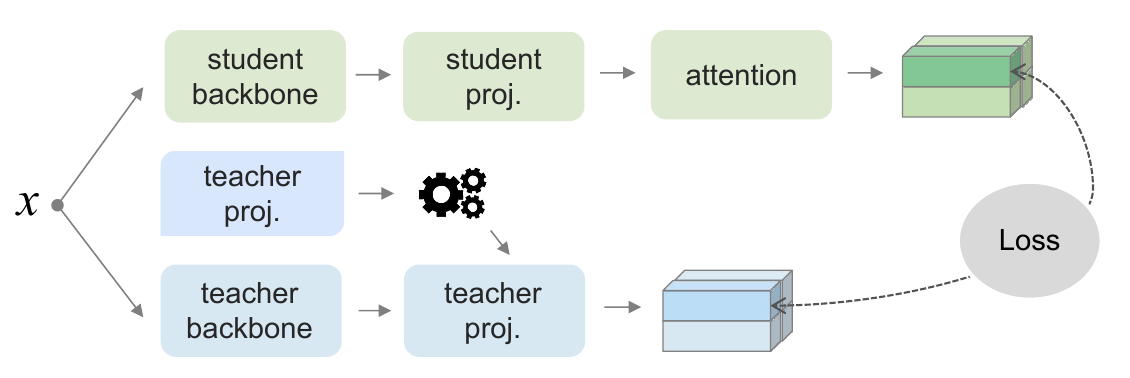}}
        \ffigbox[\FBwidth]{\caption{Workflows of Teacher's Projection Head\label{fig:spatial_adaptor}}}
        {\includegraphics[width=.47\textwidth]{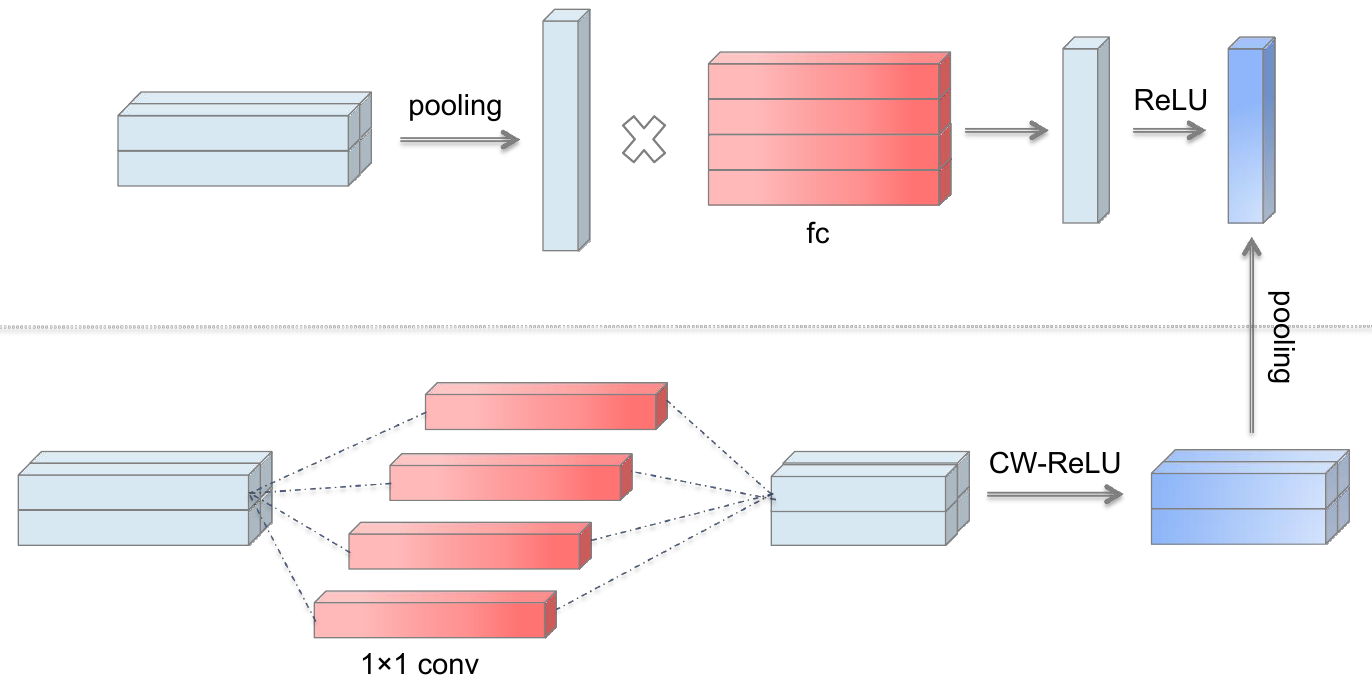}}
        \end{subfloatrow}
     \vspace{-.1em}
     \caption{(a) is the specific architecture of Pixel-Wise Contrastive Distillation. Before distillation, the original teacher's projection head is modified by SpatialAdaptor (represented by \faGears\space in the figure). Distillation loss is the average contrastive loss computed over all corresponding pixel pairs of the student and the teacher. (b) depicts the workflows of teacher's projection head before (top half) and after (bottom half) using SpatialAdaptor. The pooling layer on the far right is used for demonstrating the invariability of SpatialAdaptor. Best viewed in color.}
     \vspace{-.5em}}
    
\end{figure*}

Second, it is not straightforward to distill pixel-level knowledge from the commonly adopted teachers pre-trained by image-level SSL methods \cite{moco,simclr,byol,swav}. These teachers tend to project images into vectorized features, thus losing the spatial information which is \textit{indispensable} to pixel-to-pixel distillation in our PCD. An intuitive practice to circumvent this issue would be to remove the well-trained projection/prediction head (a non-linear MLP attached to the backbone) and the global pooling layer. The distillation loss will be computed over the feature maps output by the backbone. However, we experimentally show the ineffectiveness of such a simplistic approach, implying that the projection/prediction head contains nonnegligible knowledge for pixel-level distillation. Towards the goal of leveraging this knowledge, we introduce a \textit{SpatialAdaptor} to adapt the projection/prediction head used for encoding vectorized features to processing 2D feature maps while not changing the distribution of the output features. Though conceptually simple, the SpatialAdaptor is of great significance to pixel-level distillation. 


Last, small models are innately weak in capturing information from the regions of large spans due to their smaller effective receptive fields (ERF) \cite{erf}. This natural deficiency prevents students from further imitating teachers at pixel-level. In this case, we append a multi-head self-attention (MHSA) module \cite{transformer} to the student model, which \textit{explicitly} relates the pixels within the student's output feature maps. The addition of the MHSA module helps slightly enlarge the ERF of the small model, and consequently improves its transferring results. The MHSA will be deprecated in the fine-tuning phase. We think such gain without pain is very helpful for pre-training. We refer to \cref{fig:pcd_architecture} for a detailed depiction of PCD. 



We comprehensively evaluate the effectiveness of PCD on several typical dense prediction tasks. PCD \textit{surpasses} state-of-the-art results across all downstream tasks. Our results demonstrate the nontrivial advantages of PCD over competitive SSL methods designed for dense prediction tasks and previous image-level self-supervised distillation methods. Under the linear probing protocol, a \mbox{ResNet-18} distilled by PCD attains $65.1\%$ top-1 accuracy on ImageNet. These results highlight the superiority of pixel-level supervision signal for self-supervised distillation. 

We also find that PCD is robust to the choices of pre-trained teacher models and works well with various student backbone architectures. Students of larger backbones can compete with or even exceed the teacher on certain tasks, revealing an encouraging path for pre-training. These findings carry implications for future research and we hope our work inspires further investigation into self-supervised pre-training with small models.





\section{Related Work}
\label{sec:related_work}

\noindent \textbf{Pixel/region-level self-supervised learning} aims to learn competitive representations specialized for dense prediction tasks. Following the philosophy of contrasting pixel/region-level features from different augmented views, these methods develop various rules to find the positive pairs. 

Intuitive methods \cite{vader,scrl,resim,pixpro,instance_loc,slot_con} record the offsets and the scaling factors induced by geometric transformations (\eg, cropping, resizing, and flipping) to locate the positive pairs of pixels/regions from different augmented views. In \cite{detcon,point_level_region_contrast}, all pixels or regions within the original image are classified into some appropriate categories by a heuristic way or some unsupervised semantic segmentation methods. Any two pixels or regions from the same category form a positive pair. SoCo \cite{soco} and ORL \cite{orl} utilize the selective search \cite{selective_search} to identify numerous regions containing a single object and perform region-level contrastive learning based on these regions. DenseCL \cite{densecl} and Self-EMD \cite{selfemd} pair the pixels of feature maps from different views according to some certain rules, \eg, minimizing the cosine distances between pixels or finding the matching set with minimum earth mover's distance \cite{emd}. 

Our PCD does not rely on sophisticated rules or preparations to pair pixels or regions. Instead, we directly contrast the feature maps output by the student and the teacher from the \textit{same} view of an image, decoupling the requirement for delicate augmentation policies from the design of pre-training framework. \\

\noindent \textbf{Feature-based knowledge distillation} transfers knowledge by matching the intermediate features of students and teachers, which are often not comparable due to the difference in shapes, \ie, the number of channels and the spatial size. It is a common practice to reshape students' features to have the same shape as teachers' by a learnable module \cite{fitnets}. Some works \cite{attention_transfer,factor_transfer,overhaul,knowledge_review} transform both students' and teachers' features into tensors of the same shape. In PCD, shape alignment (especially with regard to the number of channels) is achieved by a non-linear projection head, which is a widely recognized technique in SSL for enhancing the quality of learned representations \cite{simclr,mocov2,byol}. Additionally, in cases where the student and the teacher have feature maps of different spatial sizes, we employ a simple interpolation to complete the necessary alignment. \\

\noindent \textbf{Self-supervised distillation} transfers knowledge in a self-supervised learning fashion. CompRess \cite{compress} and SEED \cite{seed} propose to minimize the feature similarity distributions between students and teachers. DoGo \cite{distill_on_the_go} and DisCo \cite{disco} add a distillation branch for easing the optimization problem of small models during self-supervised pre-training. Previous works train students to classify images \cite{cluster_fit,boosting_ssl_via_kt} or minimize the intra-group distances \cite{bingo} based on the clusterings generated by teachers. \cite{distill_on_the_fly} simultaneously trains teachers and teacher from scratch. Students are guided by teachers' on-the-fly clustering results. With these methods, the notorious problem that small models pre-trained by SSL methods face performance degradation has been partially solved. Our PCD is proposed to address the unsolved part---to improve the transferring results on dense prediction tasks.



\section{Method}
\subsection{Pixel-Wise Contrastive Distillation}
\label{sec:pcd}

Unlike general knowledge distillation methods in supervised learning, self-supervised distillation does not involve with labeled data. The supervision only comes from teachers, yielding task-agnostic students that can be fine-tuned on various downstream tasks. \\

\noindent \textbf{Image-level self-supervised distillation.} Though varying dramatically in the specific rules of computing distillation loss, current self-supervised knowledge distillation methods \cite{cluster_fit,boosting_ssl_via_kt,compress,seed,distill_on_the_go,disco,bingo,distill_on_the_fly} share one thing in common---the supervision signals are all at image-level. Below, we describe a general formulation for these methods. 

An input image is fed to the student's and the teacher's backbone, generating feature maps, $\mathbf{s} \in \mathbb{R}^{C_{s} \x H \x W}$ and $\mathbf{t} \in \mathbb{R}^{C_{t} \x H \x W}$, respectively. $C_{s}$ and $C_{t}$ are the number of channels. $H$ and $W$ are the spatial sizes\footnote{It is reasonable to assume $\mathbf{s}$ and $\mathbf{t}$ have the same spatial size owing to the popular 32-stride design of convolutional network architectures. We also discuss about the situation where the student and the teacher produce features with different spatial sizedh.}. These feature maps are then global average/max pooled into vectorized features. The distillation loss $\mathcal{L}$ with respect to a single input image is defined by: 

\begin{equation}
    \mathcal{L} = \mathcal{L}\left( \phi \left(\mathbf{s} \right), \phi \left(\mathbf{t} \right) \right),
\end{equation}

\noindent where $\phi\left(\cdot\right)$ is the global pooling layer. Note that $\mathcal{L}\left(\cdot, \cdot\right)$ is a function composition \footnote{In CompRess \cite{compress}, for example, $\mathcal{L}$ is equivalent to first estimating the similarity distributions of the student and the teacher, then computing the KL divergence among them.}, but not a simple loss function like $\ell2$ distance. We let $\phi$ be global average pooling for the ease of analysis, i.e., $\phi\left(\mathbf{x}\right)=\frac{1}{HW}\sum_{i}\mathbf{x}_{i}$. Here $i=\left(i_H, i_W\right)$ is a 2-tuple indexing the $(i_{H}, i_{W})$-th pixel of feature maps.

Given this unified formulation, we consider the derivative with respect to the $i$-th pixel of student's feature \mbox{maps $\mathbf{s}_{i}$}. By chain rule, we have:

\begin{equation} \label{eq:grad}
    \frac{\partial \mathcal{L}}{\partial \mathbf{s}_{i}} = \frac{\partial \mathcal{L}}{\partial \phi} \frac{\partial \phi}{\partial \mathbf{s}_{i}} = \frac{1}{HW} \frac{\partial \mathcal{L}}{\partial \phi}.
\end{equation}

\noindent It is obvious to see that $\frac{\partial \mathcal{L}}{\partial \mathbf{s}_{i}}$ is a term \textit{independent} from the position of the pixel. In other words, teacher's guidance is not detailed at pixel-level. There probably exists huge disparity between student's and teacher's pixel-level features. Consequently, it is far beyond reach for students (small models) to inherit the competitive pixel-level knowledge possessed by teachers. We argue that this may be the very reason of students' imbalanced performance on classification and dense prediction tasks. \\

\noindent \textbf{Pixel-Wise Contrastive Distillation.} Motivated by the above analysis, we propose a simple \textit{pixel-level} self-supervised distillation framework, Pixel-Wise Contrastive Distillation (PCD). Our PCD transfers knowledge by attracting the positive pairs of pixels from students and teachers and repulsing the negative pairs. 

Different from augmentation-invariant representation learning \cite{cmc,moco,pirl,simclr,pixpro,densecl}, the positive pairs of PCD are from the \textit{corresponding} pixels of student's and teacher's output feature maps for the same image, \ie, $\left(\mathbf{s}_{i}, \mathbf{t}_{i}\right)$. Negative samples $\{\mathbf{n}^{k} | k=1, ..., K\}$ are stored in a queue in conformity with MoCo \cite{moco}. For an input image, we optimize the average contrastive loss of all output pixels:

\begin{equation}
    \label{eq:pcd}
    \mathcal{L}\left(\mathbf{s}, \mathbf{t}\right) = \frac{1}{HW} \sum_{i} \ell\left(\mathbf{s}_{i}, \mathbf{t}_{i}, \{\mathbf{n}^{k}\}\right),
\end{equation}

\noindent where $\ell$ stands for the contrastive loss function. We do not directly contrast $\mathbf{s}_{i}$ and $\mathbf{t}_{i}$ for they may have different dimensions (\ie, the numbers of channels). Following \mbox{SimCLR} \cite{simclr}, we append a projection head (a non-linear MLP $\varphi$ parameterized by $\theta$) to student's backbone. Details of the projection head $\varphi\left(\cdot|\theta\right)$ will be given in \cref{sec:baseline_settings}. The projection head $\varphi$ serves as aligning $\mathbf{s}_{i}$ and $\mathbf{t}_{i}$ in terms of dimension, and will be removed once training is accomplished. We denote the projected output $\varphi\left(\mathbf{s}_{i} | \theta\right)$ as $\mathbf{s}_{i}^{\ast}$ for short. The concrete form of $\ell$ is:

\begin{equation}
    \label{contrastive_loss}
    \ell = -\log{\frac{\exp{\left({\mathbf{s}_{i}^{\ast}}^{\mathsf{T}} \mathbf{t}_{i} / \tau\right)}}{\exp{\left({\mathbf{s}_{i}^{\ast}}^{\mathsf{T}} \mathbf{t}_{i} / \tau\right)} + \sum_{k}^{K} \exp{\left({\mathbf{s}_{i}^{\ast}}^{\mathsf{T}} \mathbf{n}^{k} / \tau\right)}}},
\end{equation}

\noindent where $\tau$ is a temperature hyper-parameter. After back-propagation, teacher's feature maps $\mathbf{t}$ will be global pooled, $\ell2$-normalized, and enqueued as a negative sample used for subsequent iterations. Here, we omit the $\ell2$-normalization applied to $\mathbf{s}_{i}^{\ast}$ and $\mathbf{t}_{i}$. So the inner product ${\mathbf{s}_{i}^{\ast}}^{\mathsf{T}}\mathbf{t}_{i}$ equals to the cosine distance.

It is worth noting that PCD does not require $\mathbf{s}$ and $\mathbf{t}$ to have the same spatial size. In cases of $\mathbf{s}$ and $\mathbf{t}$ having mismatched spatial sizes, we perform bilinear interpolation on $\mathbf{t}$ to match the spatial size of $\mathbf{s}$. Further discussions are in \cref{sec:ablation_exps}. \\

\noindent \textbf{SpatialAdaptor.} For the models pre-trained by image-level SSL methods (\eg, \cite{moco,simclr,byol,swav}), their projection/prediction heads (henceforth projection head for simplicity) are the stacked fully-connected (fc) layers with batch normalization (BN) layers and ReLU in between. These projection heads only take the global pooled features as inputs. If adopting these models as teachers (a common practice in previous self-supervised distillation methods), one has to \textit{remove} the global pooling layer and the projection heads to be compatible with our PCD. However, removing the well-trained projection heads will break the \textit{integrity} of teachers. Such removal incurs knowledge loss, bringing in sub-optimal results for transfer learning. We will empirically verify this in \cref{sec:ablation_exps}. 

To meet the demand of utilizing the fruitful knowledge of the projection heads, we propose a \textit{SpatialAdaptor} to adapt the projection heads to processing 2D inputs. We next discuss a simple case where the projection head only contains a fc layer ($f$) and ReLU ($\sigma$), to demonstrate how the SpatialAdaptor works. Before using the SpatialAdaptor, the feature maps $\mathbf{t}$ output by teacher's backbone will be global average pooled and fed to the projection head:

\begin{align}
    \label{eq:fc_to_conv}
    \sigma\left(f\left(\phi\left(\mathbf{t}\right)\right)\right) &= \sigma\left(f\left( \frac{1}{HW} \sum_{i} \mathbf{t}_{i} \right)\right) \notag \\
    &= \sigma\left(\frac{1}{HW}\sum_{i} f\left(\mathbf{t}_{i}\right)\right).
\end{align}

\noindent The second equality holds being a consequence of $f$'s linearity. By interchanging $f$ and $\phi$, $f$ now acts on pixels rather than vectorized features. It follows that $f$ can be \textit{reformulated} into a $1\x1$ convolution (conv) layer with the stride of $1$. This simple case does not consider the existence of any BN layer since fc and BN layer together can be fused into a linear function and represented by $f$.

Furthermore, we present the Channel-Wise ReLU (CW-ReLU) that masks out the channels whose mean are negative. Let $\sigma^{\ast}$ denote CW-ReLU. By the definition, we have: 

\begin{equation}
    \label{eq:cw_relu}
    \sigma\left( \phi\left( \mathbf{x} \right) \right) = \phi \left( \sigma^{\ast}\left(\mathbf{x}\right)\right).
\end{equation}

\noindent \cref{eq:cw_relu} means the global pooling layer and the activation function now are ``\textit{commutative}''. Combining \cref{eq:fc_to_conv} with this commutative property, we can interchange the global pooling layer and the activation function:

\begin{equation}
    \label{eq:spatial_adaptor}
    \sigma\left(f\left(\phi\left(\mathbf{t}\right)\right)\right) = \phi\left( \sigma^{\ast} \left( f\left( \mathbf{t} \right) \right) \right).
\end{equation} 

\noindent Note that $\phi$ in the right-hand side of \cref{eq:spatial_adaptor} is only used for illustrating the \textit{invariability} of the SpatialAdaptor. It will be omitted in actual use for maintaining spatial information. 

Thus far, we have shown how a projection head composed of a fc layer and ReLU is adapted by the SpatialAdaptor to processing 2D feature maps while not changing the feature distribution (\cref{fig:spatial_adaptor}). As such, the integrity of teachers is maintained, and our PCD is made compatible with the teachers pre-trained by various SSL methods. Though the case discussed above being so simple, the SpatialAdaptor can be easily generalized to the situation where the projection head is more complex (\ie, stacked with more fc layers and ReLU). 


We are aware that the models pre-trained by pixel-level SSL methods \cite{pixpro,densecl} have the projection heads processing 2D inputs. But this does \textit{not} deprive the significance of the SpatialAdaptor, because we do not want too much constraints on teacher's pre-training methods. Such accessibility can also be seen as a strength of our PCD. \\

\noindent \textbf{Multi-head self-attention.} Apart from the granularity of distillation signals, the \textit{intrinsic properties} (\eg, limited capacity and smaller receptive field) of the students also play a role in self-supervised distillation. Consider the effective receptive field (ERF) \cite{erf} for an example. ERF measures how much each pixel contributes to the final prediction and has been proven to be closely related to the performance of abundant computer vision tasks \cite{deeplabv3,octconv}. Intuitively, models with larger ERF are able to capture information from a bigger area of image, leading to more robust and reliable predictions. 

\begin{figure}[t]
    \centering
    \includegraphics[width=\textwidth]{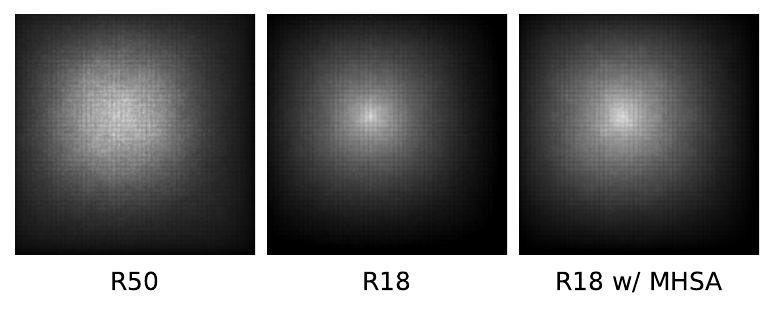}
    \caption{\textbf{Effective receptive field}. R50 and R18 are ResNet-50 and ResNet-18 respectively. \mbox{R18 w/ MHSA} stands for ResNet-18 enhanced by a multi-head self-attention module.}
    \label{fig:mhsa}
    \vspace{-.5em}
\end{figure}

According to \cite{erf}, we draw the ERFs of ResNet-50 and ResNet-18 \cite{resnet} in \cref{fig:mhsa}. We can see a clear contrast that ResNet-50 has larger ERF (larger bright region) than ResNet-18. Therefore, it is \textit{unrealistic} to expect small models to perfectly capture information from the regions of large spans like teachers do without outside help. 

The definition of ERF indicates allowing more pixels to participate in predictions helps enlarge ERF. From this perspective, we can enhance the student model by \textit{explicitly} relating all pixels right before making predictions. This is made possible by a multi-head self-attention (MHSA) module \cite{transformer}. We introduce a MHSA module between student's projection head and contrastive loss. Information from different pixels are aggregated together to make a more robust prediction. This module only induces a handful of memory and computation cost during pre-training, and does not influence the fine-tuning phase. Equipped with the MHSA module, the ERF of ResNet-18 is slightly enlarged (shown in the third picture of \cref{fig:mhsa}). 

\subsection{Baseline Settings}
\label{sec:baseline_settings}

In this section, we provide the necessary details for implementing PCD. \\

\noindent \textbf{Teacher.} In most of our experiments, we adopt the \mbox{ResNet-50} pre-trained by \mbox{MoCo v3}\footnote{The checkpoint can be found in \url{https://github.com/facebookresearch/moco-v3/blob/main/CONFIG.md}} \cite{mocov3} for 1000 epochs as the teacher. We use the momentum updated branch (\ie, \texttt{momentum\_encoder} of the checkpoint) of \mbox{MoCo v3} model. The projection head has the following structure: FC--BN--ReLU--FC--BN. The last BN has no affine parameters. We drop it because its static normalization statistics slow down the convergence. The global pooling layer of the backbone is removed. Modified by the SpatialAdaptor, the structure of the projection head becomes Conv--BN--CW-ReLU--Conv. The hidden and output dimension of the projection head is $4096$ and $256$ respectively. The shape of output feature maps is $256 \x 7 \x 7$ (channels $\x$ spatial size). Other choices of teacher models are discussed in \cref{sec:ablation_exps}. \\

\noindent \textbf{Student.} By default, we use a ResNet-18 as the student's backbone. It is followed by a projection head ($\varphi$ mentioned in \cref{sec:pcd}). We follow the asymmetric structure design in \cite{byol,simsiam,mocov3} to instantiate $\varphi$ with two consecutive MLPs. The MLPs are structurally similar to the teacher's projection head that is modified by the SpatialAdaptor, except that the activation function is ReLU. The hidden dimension of the student is equal to that of the teacher. We append a MHSA module \cite{transformer} without position encoding \cite{cvt} to the end of $\varphi$. The MHSA has $8$ heads, each with an embedding dimension of $64$. Each pixel of the input feature maps is regarded as a token. The attention values of all heads are concatenated and then projected by a $1 \x 1$ conv layer to match with the input features in dimension. The output of the MHSA module is used for computing the contrastive distillation loss.  Other choices of student backbone architectures are discussed in \cref{sec:ablation_exps}. \\

\noindent \textbf{Distilling.} We perform self-supervised distillation on the ImageNet \cite{imagenet} training set. The image augmentation policy is the same as that proposed in \cite{byol}, comprised of two distributions of augmentation. It generates a pair of augmented views for an input image. Our PCD computes distillation loss on each view and optimizes the mean of two losses. We observe this loss symmetrization brings about better convergence. We must point out that PCD with asymmetric loss also provides bright transfer performance (see in \cref{sec:main_results}). 

We use the LARS \cite{lars} optimizer to train for 100 epochs. The batch size is 1024. The base learning rate ($lr$) is set to be $1.0$ and scaled by the linear scaling rule \cite{linear_rule}: $lr \x \text{batch}\_\text{size} / 256$. The learning rate is linearly increased to $4.0$ for the first 10 epochs of training (warmup) and then decayed to $0$ based on the cosine schedule. We use a weight decay of $0.00001$ and a momentum of $0.9$. All biases and the affine parameters of BN layers are excluded from the weight decay. We set the temperature of contrastive loss to be $0.2$. The queue storing negative samples has a capacity of $65536$.


\section{Experiments}

\subsection{Evaluation Setup}

\begin{table*}[t]
\centering
\Large
\tablestyle{6pt}{1.2}
\begin{tabular}{l | x{36} | x{24}x{24} | x{24}x{24} | x{24}x{24} | x{24}x{24} | x{36}}
& ImageNet & \multicolumn{2}{c|}{VOC 07} & \multicolumn{2}{c|}{VOC 07+12} & \multicolumn{2}{c|}{COCO-C4} & \multicolumn{2}{c}{COCO-FPN} & CityScapes \\
& Acc & AP$_\text{50}$ & AP & AP$_\text{50}$ & AP & AP$^\text{bbox}$ & AP$^\text{mask}$ & AP$^\text{bbox}$ & AP$^\text{mask}$ & IoU\\
\shline
Teacher & 74.6 & 77.8 & 48.1 & 83.0 & 56.7 & 37.4 & 32.8 & 40.7 & 36.9 & 73.5 \\
\hline
ImageNet Supervised & \textbf{69.8} & 70.0 & 38.2 & 76.9 & 47.3 & 30.7 & 28.0 & 36.3 & 33.0 & 70.2 \\
\hline
MoCo v2 \cite{mocov2} & 48.7 & 69.7 & 40.2 & 77.5 & 49.7 & 30.7 & 27.9 & 35.0 & 31.8 & 70.4 \\
PixPro \cite{pixpro} & 41.4 & 71.5 & 42.3 & 78.5 & 51.1 & 30.9 & 28.1 & 35.8 & 32.6 & 70.3 \\
\hline
CompRess \cite{compress} & 63.9 & 71.3 & 41.2 & 78.4 & 50.4 & 31.4 & 28.4 & 35.7 & 32.4 & 70.3 \\
DisCo \cite{disco} & 63.5 & 62.5 & 30.7 & 72.6 & 40.1 & 28.2 & 25.8 & 36.0 & 32.8 & 69.3 \\
BINGO \cite{bingo} & 64.2 & 70.4 & 39.9 & 77.8 & 49.3 & 31.1 & 28.2 & 36.2 & 32.8 & 71.0 \\
\hline
\textbf{PCD}, asymmetric & 64.2 & \deh{72.7} & \deh{42.7} & \deh{79.3} & \deh{51.5} & \deh{31.9} & \deh{28.8} & \deh{37.0} & \deh{33.7} & \deh{71.6} \\
\textbf{PCD} & \deh{65.1} & \textbf{73.0} & \textbf{43.2} & \textbf{79.4} & \textbf{52.1} & \textbf{32.2} & \textbf{29.0} & \textbf{37.4} & \textbf{34.0} & \textbf{71.8} \\
\end{tabular}
\caption{\textbf{Comparing different pre-trained models}. All pre-trained models adopt ResNet-18 as backbone. ImageNet supervised pre-trained model is from the model zoo of PyTorch \cite{pytorch}. Other pre-trained models are from our reproductions built on their officially released codes. For fair comparisons, we pre-trained these models for 100 epochs. We use the ResNet-50 pre-trained by \mbox{MoCo v3} as teacher for all self-supervised distillation methods. The best results are marked in \textbf{bold}, and the second best are marked in \deh{gray} (exclusive of the teacher). 
\label{tab:main_results}
}
\vspace{-.5em}
\end{table*}

Here, we provide some background for our fine-tuning experiments. We evaluate our PCD on VOC \cite{voc} object detection, COCO \cite{coco} object detection and instance segmentation, and CityScapes \cite{cityscapes} semantic segmentation. The codebase for evaluation is Detectron2 \cite{detectron2}. We strictly follow the fine-tuning settings proposed in \cite{moco,pixpro} for fair comparisons. We also provide the linear probing accuracy on ImageNet. 

It has been a notorious problem that fine-tuning LARS-trained models with optimization hyper-parameters best selected for SGD-trained counterparts yields sub-optimal performance \cite{simsiam,revisiting_airl}. To address this, recent work \cite{revisiting_airl} proposes NormRescale to scale the norm of LARS-trained weights by a specific anchor (\eg, the norm of SGD-trained weights or a constant number). It helps the LARS-trained models fit to the optimization strategy of fine-tuning. When fine-tuning C4 or FCN backbones pre-trained by PCD, we employ this technique to multiply the weights by a constant $0.25$. Multiplying a constant is an efficient choice for not introducing extra training cost. \\

\noindent \textbf{VOC object detection.} We use a C4 backbone with Faster R-CNN \cite{frcnn} detector. We evaluate the pre-trained models under two fine-tuning settings. The first is to train on \texttt{trainval2007} set for 9k iterations, and the second is to train on \texttt{trainval07+12} set for 24k iterations. We use the same fine-tuning settings as per \cite{moco}. Fine-tuned models are evaluated on \texttt{test2007} set. For better reproducibility, we report the average AP$_{50}$ and AP over 5 runs. \\

\noindent \textbf{COCO object detection and instance segmentation.} We use two types of backbone, C4 and FPN, for fine-tuning on COCO dataset. The detector is Mask R-CNN \cite{mrcnn}. Pre-trained models are fine-tuned on \texttt{train2017} set according to the $1\x$ optimization setting (about 12 COCO epochs). We report AP$^\text{bbox}$ for object detection and AP$^\text{mask}$ for instance segmentation on \texttt{val2017} set. \\

\noindent \textbf{CityScapes semantic segmentation.} We implement a FCN-like \cite{fcn} structure based on pre-trained backbones. A newly initialized BN layer is added to the end of pre-trained backbones for helping optimization. Subsequently, we append two atrous convolutional blocks, each with a $3 \x 3$ conv layer of 256 output channels, a BN layer, and ReLU. The conv layers have stride 1, dilation 6, and padding 6. The prediction layer is a $1 \x 1$ conv layer with $19$ output channels ($19$ classes), whose outputs are then bilinearly interpolated to match the size of input images. Fine-tuning takes 90k iterations on \texttt{train\_fine} set. More detailed information of fine-tuning can be found in Appendix. We report the average IoU on \texttt{val} set over 5 runs. \\

\noindent \textbf{Linear probing in ImageNet.} We freeze the backbones pre-trained by PCD and train a linear classifier on the ImageNet training set. We use nesterov SGD to train for 100 epochs. The batch size is 1024, and the learning rate is $0.8$ (base \textit{lr} $=0.2$). The learning rate will decay to $0$ according to the cosine schedule without restart. The momentum is $0.9$, and the weight decay is $0$. We use a vanilla image augmentation policy containing random cropping, resizing to $224 \x 224$, and horizontal flipping. We report the single-crop classification accuracy on the ImageNet validation set. \\

\subsection{Main Results} \label{sec:main_results}

In \cref{tab:main_results}, we present the fine-tuning results of the following pre-training methods: supervised pre-training, SSL methods, previous competitive self-supervised distillation methods, and our PCD. We notice that PixPro \cite{pixpro} (a SSL method designed for dense prediction tasks) outperforms the self-supervised distillation methods on most tasks. This observation confirms our hypothesis that small models are \textit{difficult} to learn pixel-level knowledge from image-level pretext tasks, even with distillation signals, further justifying the necessity of our method.

Our PCD shows impressive generalization capacity: it surpasses all competitors on each dense prediction task. We achieves $37.4$ AP$^\text{bbox}$ and $34.0$ AP$^\text{mask}$ on COCO using the \mbox{Mask R-CNN} detector and the \mbox{ResNet-18-FPN} backbone, emerging as the first pre-training method exceeding the supervised pre-trained model on this benchmark. Under the linear probing protocol, our PCD also achieves decent top-1 accuracy ($65.1\%$), which makes PCD a well-rounded self-supervised distillation method. 

We notice that some competitors (\eg, supervised learning, \mbox{MoCo v2} \cite{mocov2}, and CompRess \cite{compress}) are pre-trained by asymmetric loss. Here, we provide an asymmetric variant of PCD to exclude the effect of loss symmetrization. The change is simple: we adopt the symmetric augmentation policy as per \cite{byol}, and sample one augmented view from each input image during training. Indeed, the symmetric loss endows PCD with better performance, but asymmetric variant still achieves the second best results on all tasks (marked as gray in \cref{tab:main_results}). The \textit{nontrivial} advantages of PCD against other self-supervised distillation methods have confirmed the fact that pixel-level distillation signals are the key to transferring knowledge conducive to dense prediction tasks. 

\subsection{Ablation Experiments}
\label{sec:ablation_exps}

We perform extensive ablation experiments to analyze PCD. Unless specified, we adopt the training settings mentioned in \cref{sec:baseline_settings}. \\

\begin{table}[t]
\centering
\large
\tablestyle{4pt}{1.2}
\begin{tabular}{l |x{24}x{24} | x{24}x{24} | x{36}}
& \multicolumn{2}{c|}{VOC 07+12} & \multicolumn{2}{c|}{COCO-FPN} & CityScapes \\
& AP$_\text{50}$ & AP & AP$^\text{bbox}$ & AP$^\text{mask}$ & IoU\\
\shline
image-level & 78.7 & 49.8 & 36.5 & 33.3 & 71.1 \\
pixel-level & \textbf{79.4} & \textbf{52.1} & \textbf{37.4} & \textbf{34.0} & \textbf{71.8} \\
\end{tabular}
\vspace{-.1em}
\caption{\textbf{Pixel-level \vs image-level}. We compare PCD to image-level contrastive distillation. The backbone is ResNet-18 and pre-trained for 100 epochs. The best results are marked in \textbf{bold}.
\label{tab:pixel_vs_image}
}
\vspace{-.5em}
\end{table}

\begin{table}[t]
\centering
\large
\tablestyle{4pt}{1.2}
\begin{tabular}{l |x{24}x{24} | x{24}x{24} | x{36}}
& \multicolumn{2}{c|}{VOC 07+12} & \multicolumn{2}{c|}{COCO-FPN} & CityScapes \\
& AP$_\text{50}$ & AP & AP$^\text{bbox}$ & AP$^\text{mask}$ & IoU\\
\shline
(a) & 76.0 & 48.8 & 36.5 & 33.2 & 70.1 \\
(b) & 77.9 & 50.6 & 36.8 & 33.6 & 70.7 \\
(c) & 77.3 & 50.1 & 36.3 & 33.1 & 69.6 \\
(d) & 77.8 & 50.7 & 36.6 & 33.2 & 70.1 \\
ours & \textbf{79.4} & \textbf{52.1} & \textbf{37.4} & \textbf{34.0} & \textbf{71.8} \\
\end{tabular}
\vspace{-.1em}
\caption{\textbf{Ablations on SpatialAdaptor}. We compare four variants of PCD (a-d) to examine the necessity of SpatialAdaptor. The backbone is ResNet-18 and pre-trained for 100 epochs. The best results are marked in \textbf{bold}.
\label{tab:spatial_adaptor}
}
\vspace{-.5em}
\end{table}

\noindent \textbf{Pixel-level \vs\ image-level.} To directly compare pixel-level and image-level distillation, we develop an image-level variant of PCD. Based on PCD, this variant vectorizes student's and teacher's output feature maps by an extra global average pooling layer and computes contrastive loss on these vectorized features. It has highly competitive results (\cref{tab:pixel_vs_image}) like those image-level self-supervised distillation methods in \cref{tab:main_results}, revealing the effectiveness of contrastive loss used for self-supervised distillation. But it still \textit{lags behind} the original PCD on all downstream tasks. This gap further confirms the importance of pixel-level distillation signal. \\

\begin{table}[t]
\centering
\large
\tablestyle{4pt}{1.2}
\begin{tabular}{l |x{24}x{24} | x{24}x{24} | x{36}}
& \multicolumn{2}{c|}{VOC 07+12} & \multicolumn{2}{c|}{COCO-FPN} & CityScapes \\
& AP$_\text{50}$ & AP & AP$^\text{bbox}$ & AP$^\text{mask}$ & IoU\\
\shline
w/o MHSA & 79.2 & 51.8 & 37.1 & 33.7 & 71.6 \\
extra pred. & 79.1 & 52.0 & 37.2 & 33.7 & 71.2 \\
ours & \textbf{79.4} & \textbf{52.1} & \textbf{37.4} & \textbf{34.0} & \textbf{71.8} \\
\end{tabular}
\vspace{-.1em}
\caption{\textbf{Ablations on MHSA}. We ablate the MHSA module. ``extra pred.'' stands for replacing the MHSA module by an extra prediction head. The backbone is ResNet-18 and pre-trained for 100 epochs. The best results are marked in \textbf{bold}.
\label{tab:mhsa}
}
\vspace{-.5em}
\end{table}

\begin{table*}[t]
\centering
\Large
\tablestyle{6pt}{1.2}
\begin{tabular}{l | x{54} | x{24}x{24} | x{24}x{24} | x{24}x{24} | x{24}x{24} | x{36}}
\multirow{2}{*}{backbone} & \multirow{2}{*}{\tablestyle{0pt}{0.9} \begin{tabular}{c} {pre-training} \\ {method} \end{tabular}} & \multicolumn{2}{c|}{VOC 07} & \multicolumn{2}{c|}{VOC 07+12} & \multicolumn{2}{c|}{COCO-C4} & \multicolumn{2}{c}{COCO-FPN} & CityScapes \\
& & AP$_\text{50}$ & AP & AP$_\text{50}$ & AP & AP$^\text{bbox}$ & AP$^\text{mask}$ & AP$^\text{bbox}$ & AP$^\text{mask}$ & IoU\\
\shline
ResNet-50 & SwAV & 72.1 & 41.7 & 79.0 & 51.2 & 31.9 & 28.8 & 37.1 & 33.7 & 70.7 \\
ResNet-50 & BYOL & 71.7 & 42.6 & 78.9 & 51.5 & 32.0 & 29.0 & 37.0 & 33.7 & 71.0 \\
ResNet-50 & Barlow Twins & 70.7 & 41.6 & 78.4 & 50.7 & 31.3 & 28.4 & 35.9 & 32.7 & 70.9 \\
ViT-Base & MoCo v3 & 70.6 & 41.7 & 78.1 & 51.1 & 31.3 & 28.5 & 36.6 & 33.6 & 71.4 \\

\end{tabular}
\vspace{-.1em}
\caption{\textbf{Fine-tuning results of PCD with different teachers}. The student backbone is ResNet-18. All teacher models are from the officially released checkpoints. We refer to Appendix for more details about these teachers.
\label{tab:different_teachers}
}
\vspace{-.5em}
\end{table*}

\begin{table*}[t]
\centering
\Large
\tablestyle{6pt}{1.2}
\begin{tabular}{l | x{48} | x{24}x{24} | x{24}x{24} | x{24}x{24} | x{24}x{24} | x{36}}
& pre-training & \multicolumn{2}{c|}{VOC 07} & \multicolumn{2}{c|}{VOC 07+12} & \multicolumn{2}{c|}{COCO-C4} & \multicolumn{2}{c}{COCO-FPN} & CityScapes \\
& method & AP$_\text{50}$ & AP & AP$_\text{50}$ & AP & AP$^\text{bbox}$ & AP$^\text{mask}$ & AP$^\text{bbox}$ & AP$^\text{mask}$ & IoU \\
\shline
\multirow{2}{*}{ResNet-34} & supervised & 74.8 & 45.0 & 81.0 & 55.0 & 37.7 & 32.9 & 39.4 & 35.6 & 71.9 \\
 & PCD & \textbf{76.3}  & \textbf{49.1} & \textbf{82.0} & \textbf{58.2} & \textbf{38.2} & \textbf{33.7} & \textbf{40.9} & \textbf{37.0} & \textbf{73.2} \\
\hline
\multirow{2}{*}{ResNet-50} & supervised & 75.2 & 44.5 & 81.5 & 54.1 & 38.2 & 33.5 & 40.2 & 36.3 & 72.3 \\
 & PCD & \textbf{77.0} & \textbf{49.0} & \textbf{82.8} & \textbf{57.7} & \textbf{40.1} & \textbf{34.9} & \textbf{42.4} & \textbf{38.1} & \textbf{73.3} \\
\hline
\multirow{2}{*}{MobileNet v3} & supervised & 67.9 & 36.7 & 76.2 & 46.4 & 30.6 & 27.9 & 35.8 & 32.7 & 68.3 \\
 & PCD & \textbf{73.9} & \textbf{40.8} & \textbf{79.3} & \textbf{49.9} & \textbf{32.8} & \textbf{29.6} & \textbf{37.7} & \textbf{34.3} & \textbf{69.0} \\

\end{tabular}
\vspace{-.1em}
\caption{\textbf{Fine-tuning results of PCD with different student backbones}. The teacher used for PCD is ResNet-50 pre-trained by \mbox{MoCo v3}. The student backbones are ResNet-18, ResNet-34, ResNet-50, and \mbox{MobileNet v3 (Large)}. We also fine-tune the supervised pre-trained counterparts for contrast. We refer to Appendix for more implementation details. The best results for each backbone are marked as \textbf{bold}.
\label{tab:different_students}
}
\vspace{-.5em}
\end{table*}

\noindent \textbf{Ablation on SpatialAdaptor.} We examine the necessity of the SpatialAdaptor for learning competitive representations. Without resorting to the SpatialAdaptor, we remove teacher's projection head (along with the global pooling layer) and simply use the feature maps output by teacher's backbone to compute contrastive loss (variant (a) in \cref{tab:spatial_adaptor}). The evaluation metrics AP$_{50}$ and AP on VOC are rather low. This variant overlooks the fact that teacher's feature maps have numerous zeros (the characteristic of ReLU) while student's feature maps do not. Contrasting two features from different distributions naturally leads to sub-optimal results. 

For more reasonable comparisons, we introduce two more variants extended from variant (a). Variant (b) adds ReLU after the MHSA module. Variant (c) removes the MHSA module and adds ReLU after student's projection head. Overall, these two variants (\cref{tab:spatial_adaptor} (b-c)) are still significantly worse than PCD. And they are no better than the image-level self-supervised distillation methods in \cref{tab:main_results} and \cref{tab:pixel_vs_image}. We argue that preserving the integrity of teachers (with the help of the SpatialAdaptor) is of vital importance to pixel-level distillation. Otherwise, it will notably lower the quality of learned representations. 

Additionally, we study the impact of invariability of SpatialAdaptor by replacing CW-ReLU with ReLU (variant (d)). We observe that keeping the distribution of teacher's learned features unchanged has massive gains on dense prediction tasks (\cref{tab:spatial_adaptor} (d)). In sum, the SpatialAdaptor is an essential component of PCD, enabling more \textit{informative} pixel-wise distillation from teachers pre-trained by image-level SSL methods. \\

\noindent \textbf{Ablation on multi-head self-attention.} We ablate the MHSA module in \cref{tab:mhsa}. PCD without the MHSA module meets slight performance drop. A plausible explanation for the positive impact of the MHSA module is that it works like a prediction head to promote the quality of learned representations \cite{byol,simsiam,mocov3}. We thus supersede the MHSA module by an extra prediction head of roughly the same number of parameters. This substitution does not bring any improvement (\cref{tab:mhsa}), suggesting that explicitly relating pixels is useful for PCD. The MHSA module only adds a small computational overhead to the pre-training phase, but it consistently benefits to various downstream tasks. We therefore regard it as a necessary part to PCD. \\

\noindent \textbf{Different teachers.} Both \mbox{MoCo v3} and PCD are trained with contrastive loss. To \textit{exclude} the positive or negative effect induced by optimizing the same type of loss, we consider using teachers pre-trained by SwAV \cite{swav}, BYOL \cite{byol}, and \mbox{Barlow Twins} \cite{barlow_twins}. The fine-tuning results in \cref{tab:different_teachers} show that these teachers can also inspire favorable representations. The effectiveness of PCD is not strictly correlated to the teacher model pre-trained by \mbox{MoCo v3}. It can be concluded that PCD is \textit{robust} to the choices of teacher models. 

Beyond typical convolutional architectures, we try using ViT \cite{vit} as the teacher to study the effect of cross-architectures distillation. There is an innate obstacle for ResNet-18 to imitate ViT at pixel-level, since they differ in the resolutions\footnote{ViT-Base has $14 \x 14$ output patches for a $224 \x 224$ input, and we treat each patch as a pixel.} of output feature maps. Therefore, we downsample the output of ViT by a $2 \x 2$ average pooling layer with a stride of 2. Another solution would be to employ strided convolution in the projection head of student model. We leave it to be a topic of future research. It leads to acceptable results on dense prediction tasks, whereas unsatisfying linear top-1 accuracy ($57.6\%$) on ImageNet. We believe cross-architectures distillation (between CNNs and transformers) is a noteworthy problem for future research. \\

\noindent \textbf{Different students.} We investigate the effectiveness of PCD on different student backbones: ResNet-34, ResNet-50, and \mbox{MobileNet v3 (Large)} \cite{mobilenetv3}. Compared to supervised pre-training, our PCD consistently outperforms on all backbones (\cref{tab:different_students}). A clear \textit{trend} is that backbones with larger capacity (from \mbox{ResNet-18} to \mbox{ResNet-50}) have better transfer performance. A distilled \mbox{ResNet-34} or \mbox{ResNet-50} can rival or even beat the teacher (referred to \cref{tab:main_results}) on some downstream tasks, marking the practicability of our PCD. 


\section{Conclusion}

In this paper, we study the notorious problem that small models pre-trained by SSL methods faces performance degradation on downstream tasks, especially on dense prediction tasks. We find it difficult for small models to learn pixel-level knowledge from image-level pretext tasks, even with distillation signals. To address this problem, we propose a simple but effective self-supervised distillation framework friendly to dense prediction tasks. Given the remarkable performance of PCD, we believe it points out a practical solution to pre-training small models in a self-supervised fashion.

\section{Acknowledgements}

Junqiang Huang would like to thank his mother \mbox{Caixia Xu} and his wife \mbox{Yuwei Lin} for their support and help along the way. We are grateful for the selfless assistance offered by our friend, Chengpeng Chen.

{\small
\definecolor{Gray}{gray}{0.5}
\renewcommand\UrlFont{\color{Gray}\rmfamily}
\bibliographystyle{ieee_fullname}
\bibliography{ref}

\begin{thebibliography}{10}\itemsep=-1pt

\bibitem{compress}
Soroush Abbasi~Koohpayegani, Ajinkya Tejankar, and Hamed Pirsiavash.
\newblock Compress: Self-supervised learning by compressing representations.
\newblock {\em Advances in Neural Information Processing Systems},
  33:12980--12992, 2020.

\bibitem{need_to_be_deep}
Jimmy Ba and Rich Caruana.
\newblock Do deep nets really need to be deep?
\newblock {\em Advances in neural information processing systems}, 27, 2014.

\bibitem{point_level_region_contrast}
Yutong Bai, Xinlei Chen, Alexander Kirillov, Alan Yuille, and Alexander~C Berg.
\newblock Point-level region contrast for object detection pre-training.
\newblock In {\em Proceedings of the IEEE/CVF Conference on Computer Vision and
  Pattern Recognition}, pages 16061--16070, 2022.

\bibitem{distill_on_the_go}
Prashant Bhat, Elahe Arani, and Bahram Zonooz.
\newblock Distill on the go: Online knowledge distillation in self-supervised
  learning.
\newblock In {\em Proceedings of the IEEE/CVF Conference on Computer Vision and
  Pattern Recognition}, pages 2678--2687, 2021.

\bibitem{model_compression}
Cristian Buciluǎ, Rich Caruana, and Alexandru Niculescu-Mizil.
\newblock Model compression.
\newblock In {\em Proceedings of the 12th ACM SIGKDD international conference
  on Knowledge discovery and data mining}, pages 535--541, 2006.

\bibitem{swav}
Mathilde Caron, Ishan Misra, Julien Mairal, Priya Goyal, Piotr Bojanowski, and
  Armand Joulin.
\newblock Unsupervised learning of visual features by contrasting cluster
  assignments.
\newblock {\em Advances in Neural Information Processing Systems},
  33:9912--9924, 2020.

\bibitem{deeplabv3}
Liang-Chieh Chen, George Papandreou, Florian Schroff, and Hartwig Adam.
\newblock Rethinking atrous convolution for semantic image segmentation.
\newblock {\em arXiv preprint arXiv:1706.05587}, 2017.

\bibitem{knowledge_review}
Pengguang Chen, Shu Liu, Hengshuang Zhao, and Jiaya Jia.
\newblock Distilling knowledge via knowledge review.
\newblock In {\em Proceedings of the IEEE/CVF Conference on Computer Vision and
  Pattern Recognition}, pages 5008--5017, 2021.

\bibitem{simclr}
Ting Chen, Simon Kornblith, Mohammad Norouzi, and Geoffrey Hinton.
\newblock A simple framework for contrastive learning of visual
  representations.
\newblock In {\em International conference on machine learning}, pages
  1597--1607. PMLR, 2020.

\bibitem{mocov2}
Xinlei Chen, Haoqi Fan, Ross Girshick, and Kaiming He.
\newblock Improved baselines with momentum contrastive learning.
\newblock {\em arXiv preprint arXiv:2003.04297}, 2020.

\bibitem{simsiam}
Xinlei Chen and Kaiming He.
\newblock Exploring simple siamese representation learning.
\newblock In {\em Proceedings of the IEEE/CVF Conference on Computer Vision and
  Pattern Recognition}, pages 15750--15758, 2021.

\bibitem{mocov3}
Xinlei Chen, Saining Xie, and Kaiming He.
\newblock An empirical study of training self-supervised vision transformers.
\newblock In {\em Proceedings of the IEEE/CVF International Conference on
  Computer Vision}, pages 9640--9649, 2021.

\bibitem{octconv}
Yunpeng Chen, Haoqi Fan, Bing Xu, Zhicheng Yan, Yannis Kalantidis, Marcus
  Rohrbach, Shuicheng Yan, and Jiashi Feng.
\newblock Drop an octave: Reducing spatial redundancy in convolutional neural
  networks with octave convolution.
\newblock In {\em Proceedings of the IEEE/CVF International Conference on
  Computer Vision}, pages 3435--3444, 2019.

\bibitem{distill_on_the_fly}
Hee~Min Choi, Hyoa Kang, and Dokwan Oh.
\newblock Unsupervised representation transfer for small networks: I believe i
  can distill on-the-fly.
\newblock {\em Advances in Neural Information Processing Systems},
  34:24645--24658, 2021.

\bibitem{cityscapes}
Marius Cordts, Mohamed Omran, Sebastian Ramos, Timo Rehfeld, Markus Enzweiler,
  Rodrigo Benenson, Uwe Franke, Stefan Roth, and Bernt Schiele.
\newblock The cityscapes dataset for semantic urban scene understanding.
\newblock In {\em Proceedings of the IEEE conference on computer vision and
  pattern recognition}, pages 3213--3223, 2016.

\bibitem{vit}
Alexey Dosovitskiy, Lucas Beyer, Alexander Kolesnikov, Dirk Weissenborn,
  Xiaohua Zhai, Thomas Unterthiner, Mostafa Dehghani, Matthias Minderer, Georg
  Heigold, Sylvain Gelly, et~al.
\newblock An image is worth 16x16 words: Transformers for image recognition at
  scale.
\newblock {\em arXiv preprint arXiv:2010.11929}, 2020.

\bibitem{voc}
Mark Everingham, SM Eslami, Luc Van~Gool, Christopher~KI Williams, John Winn,
  and Andrew Zisserman.
\newblock The pascal visual object classes challenge: A retrospective.
\newblock {\em International journal of computer vision}, 111(1):98--136, 2015.

\bibitem{seed}
Zhiyuan Fang, Jianfeng Wang, Lijuan Wang, Lei Zhang, Yezhou Yang, and Zicheng
  Liu.
\newblock Seed: Self-supervised distillation for visual representation.
\newblock {\em arXiv preprint arXiv:2101.04731}, 2021.

\bibitem{disco}
Yuting Gao, Jia-Xin Zhuang, Ke Li, Hao Cheng, Xiaowei Guo, Feiyue Huang,
  Rongrong Ji, and Xing Sun.
\newblock Disco: Remedy self-supervised learning on lightweight models with
  distilled contrastive learning.
\newblock {\em arXiv preprint arXiv:2104.09124}, 2021.

\bibitem{linear_rule}
Priya Goyal, Piotr Doll{\'a}r, Ross Girshick, Pieter Noordhuis, Lukasz
  Wesolowski, Aapo Kyrola, Andrew Tulloch, Yangqing Jia, and Kaiming He.
\newblock Accurate, large minibatch sgd: Training imagenet in 1 hour.
\newblock {\em arXiv preprint arXiv:1706.02677}, 2017.

\bibitem{byol}
Jean-Bastien Grill, Florian Strub, Florent Altch{\'e}, Corentin Tallec, Pierre
  Richemond, Elena Buchatskaya, Carl Doersch, Bernardo Avila~Pires, Zhaohan
  Guo, Mohammad Gheshlaghi~Azar, et~al.
\newblock Bootstrap your own latent-a new approach to self-supervised learning.
\newblock {\em Advances in neural information processing systems},
  33:21271--21284, 2020.

\bibitem{contrastive_learning}
Raia Hadsell, Sumit Chopra, and Yann LeCun.
\newblock Dimensionality reduction by learning an invariant mapping.
\newblock In {\em 2006 IEEE Computer Society Conference on Computer Vision and
  Pattern Recognition (CVPR'06)}, volume~2, pages 1735--1742. IEEE, 2006.

\bibitem{mae}
Kaiming He, Xinlei Chen, Saining Xie, Yanghao Li, Piotr Doll{\'a}r, and Ross
  Girshick.
\newblock Masked autoencoders are scalable vision learners.
\newblock In {\em Proceedings of the IEEE/CVF Conference on Computer Vision and
  Pattern Recognition}, pages 16000--16009, 2022.

\bibitem{moco}
Kaiming He, Haoqi Fan, Yuxin Wu, Saining Xie, and Ross Girshick.
\newblock Momentum contrast for unsupervised visual representation learning.
\newblock In {\em Proceedings of the IEEE/CVF conference on computer vision and
  pattern recognition}, pages 9729--9738, 2020.

\bibitem{mrcnn}
Kaiming He, Georgia Gkioxari, Piotr Doll{\'a}r, and Ross Girshick.
\newblock Mask r-cnn.
\newblock In {\em Proceedings of the IEEE international conference on computer
  vision}, pages 2961--2969, 2017.

\bibitem{resnet}
Kaiming He, Xiangyu Zhang, Shaoqing Ren, and Jian Sun.
\newblock Deep residual learning for image recognition.
\newblock In {\em Proceedings of the IEEE conference on computer vision and
  pattern recognition}, pages 770--778, 2016.

\bibitem{detcon}
Olivier~J H{\'e}naff, Skanda Koppula, Jean-Baptiste Alayrac, Aaron Van~den
  Oord, Oriol Vinyals, and Jo{\~a}o Carreira.
\newblock Efficient visual pretraining with contrastive detection.
\newblock In {\em Proceedings of the IEEE/CVF International Conference on
  Computer Vision}, pages 10086--10096, 2021.

\bibitem{overhaul}
Byeongho Heo, Jeesoo Kim, Sangdoo Yun, Hyojin Park, Nojun Kwak, and Jin~Young
  Choi.
\newblock A comprehensive overhaul of feature distillation.
\newblock In {\em Proceedings of the IEEE/CVF International Conference on
  Computer Vision}, pages 1921--1930, 2019.

\bibitem{hinton_kd}
Geoffrey Hinton, Oriol Vinyals, Jeff Dean, et~al.
\newblock Distilling the knowledge in a neural network.
\newblock {\em arXiv preprint arXiv:1503.02531}, 2(7), 2015.

\bibitem{emd}
Frank~L Hitchcock.
\newblock The distribution of a product from several sources to numerous
  localities.
\newblock {\em Journal of mathematics and physics}, 20(1-4):224--230, 1941.

\bibitem{mobilenetv3}
Andrew Howard, Mark Sandler, Grace Chu, Liang-Chieh Chen, Bo Chen, Mingxing
  Tan, Weijun Wang, Yukun Zhu, Ruoming Pang, Vijay Vasudevan, et~al.
\newblock Searching for mobilenetv3.
\newblock In {\em Proceedings of the IEEE/CVF international conference on
  computer vision}, pages 1314--1324, 2019.

\bibitem{revisiting_airl}
Junqiang Huang, Xiangwen Kong, and Xiangyu Zhang.
\newblock Revisiting the critical factors of augmentation-invariant
  representation learning.
\newblock In {\em European Conference on Computer Vision}, pages 42--58.
  Springer, 2022.

\bibitem{factor_transfer}
Jangho Kim, SeongUk Park, and Nojun Kwak.
\newblock Paraphrasing complex network: Network compression via factor
  transfer.
\newblock {\em Advances in neural information processing systems}, 31, 2018.

\bibitem{coco}
Tsung-Yi Lin, Michael Maire, Serge Belongie, James Hays, Pietro Perona, Deva
  Ramanan, Piotr Doll{\'a}r, and C~Lawrence Zitnick.
\newblock Microsoft coco: Common objects in context.
\newblock In {\em European conference on computer vision}, pages 740--755.
  Springer, 2014.

\bibitem{selfemd}
Songtao Liu, Zeming Li, and Jian Sun.
\newblock Self-emd: Self-supervised object detection without imagenet.
\newblock {\em arXiv preprint arXiv:2011.13677}, 2020.

\bibitem{fcn}
Jonathan Long, Evan Shelhamer, and Trevor Darrell.
\newblock Fully convolutional networks for semantic segmentation.
\newblock In {\em Proceedings of the IEEE conference on computer vision and
  pattern recognition}, pages 3431--3440, 2015.

\bibitem{erf}
Wenjie Luo, Yujia Li, Raquel Urtasun, and Richard Zemel.
\newblock Understanding the effective receptive field in deep convolutional
  neural networks.
\newblock {\em Advances in neural information processing systems}, 29, 2016.

\bibitem{pirl}
Ishan Misra and Laurens van~der Maaten.
\newblock Self-supervised learning of pretext-invariant representations.
\newblock In {\em Proceedings of the IEEE/CVF Conference on Computer Vision and
  Pattern Recognition}, pages 6707--6717, 2020.

\bibitem{boosting_ssl_via_kt}
Mehdi Noroozi, Ananth Vinjimoor, Paolo Favaro, and Hamed Pirsiavash.
\newblock Boosting self-supervised learning via knowledge transfer.
\newblock In {\em Proceedings of the IEEE conference on computer vision and
  pattern recognition}, pages 9359--9367, 2018.

\bibitem{vader}
Pedro~O O~Pinheiro, Amjad Almahairi, Ryan Benmalek, Florian Golemo, and Aaron~C
  Courville.
\newblock Unsupervised learning of dense visual representations.
\newblock {\em Advances in Neural Information Processing Systems},
  33:4489--4500, 2020.

\bibitem{pytorch}
Adam Paszke, Sam Gross, Francisco Massa, Adam Lerer, James Bradbury, Gregory
  Chanan, Trevor Killeen, Zeming Lin, Natalia Gimelshein, Luca Antiga, Alban
  Desmaison, Andreas Kopf, Edward Yang, Zachary DeVito, Martin Raison, Alykhan
  Tejani, Sasank Chilamkurthy, Benoit Steiner, Lu Fang, Junjie Bai, and Soumith
  Chintala.
\newblock Pytorch: An imperative style, high-performance deep learning library.
\newblock In H. Wallach, H. Larochelle, A. Beygelzimer, F. d\textquotesingle
  Alch\'{e}-Buc, E. Fox, and R. Garnett, editors, {\em Advances in Neural
  Information Processing Systems 32}, pages 8024--8035. Curran Associates,
  Inc., 2019.

\bibitem{frcnn}
Shaoqing Ren, Kaiming He, Ross Girshick, and Jian Sun.
\newblock Faster r-cnn: Towards real-time object detection with region proposal
  networks.
\newblock {\em Advances in neural information processing systems}, 28, 2015.

\bibitem{scrl}
Byungseok Roh, Wuhyun Shin, Ildoo Kim, and Sungwoong Kim.
\newblock Spatially consistent representation learning.
\newblock In {\em Proceedings of the IEEE/CVF Conference on Computer Vision and
  Pattern Recognition}, pages 1144--1153, 2021.

\bibitem{fitnets}
Adriana Romero, Nicolas Ballas, Samira~Ebrahimi Kahou, Antoine Chassang, Carlo
  Gatta, and Yoshua Bengio.
\newblock Fitnets: Hints for thin deep nets.
\newblock {\em arXiv preprint arXiv:1412.6550}, 2014.

\bibitem{imagenet}
Olga Russakovsky, Jia Deng, Hao Su, Jonathan Krause, Sanjeev Satheesh, Sean Ma,
  Zhiheng Huang, Andrej Karpathy, Aditya Khosla, Michael Bernstein, et~al.
\newblock Imagenet large scale visual recognition challenge.
\newblock {\em International journal of computer vision}, 115(3):211--252,
  2015.

\bibitem{cmc}
Yonglong Tian, Dilip Krishnan, and Phillip Isola.
\newblock Contrastive multiview coding.
\newblock In {\em European conference on computer vision}, pages 776--794.
  Springer, 2020.

\bibitem{selective_search}
Jasper~RR Uijlings, Koen~EA Van De~Sande, Theo Gevers, and Arnold~WM Smeulders.
\newblock Selective search for object recognition.
\newblock {\em International journal of computer vision}, 104(2):154--171,
  2013.

\bibitem{access_unlabeled_data}
Ruth Urner, Shai Shalev-Shwartz, and Shai Ben-David.
\newblock Access to unlabeled data can speed up prediction time.
\newblock In {\em ICML}, 2011.

\bibitem{transformer}
Ashish Vaswani, Noam Shazeer, Niki Parmar, Jakob Uszkoreit, Llion Jones,
  Aidan~N Gomez, {\L}ukasz Kaiser, and Illia Polosukhin.
\newblock Attention is all you need.
\newblock {\em Advances in neural information processing systems}, 30, 2017.

\bibitem{densecl}
Xinlong Wang, Rufeng Zhang, Chunhua Shen, Tao Kong, and Lei Li.
\newblock Dense contrastive learning for self-supervised visual pre-training.
\newblock In {\em Proceedings of the IEEE/CVF Conference on Computer Vision and
  Pattern Recognition}, pages 3024--3033, 2021.

\bibitem{soco}
Fangyun Wei, Yue Gao, Zhirong Wu, Han Hu, and Stephen Lin.
\newblock Aligning pretraining for detection via object-level contrastive
  learning.
\newblock {\em Advances in Neural Information Processing Systems},
  34:22682--22694, 2021.

\bibitem{slot_con}
Xin Wen, Bingchen Zhao, Anlin Zheng, Xiangyu Zhang, and Xiaojuan Qi.
\newblock Self-supervised visual representation learning with semantic
  grouping.
\newblock {\em arXiv preprint arXiv:2205.15288}, 2022.

\bibitem{cvt}
Haiping Wu, Bin Xiao, Noel Codella, Mengchen Liu, Xiyang Dai, Lu Yuan, and Lei
  Zhang.
\newblock Cvt: Introducing convolutions to vision transformers.
\newblock In {\em Proceedings of the IEEE/CVF International Conference on
  Computer Vision}, pages 22--31, 2021.

\bibitem{tinyvit}
Kan Wu, Jinnian Zhang, Houwen Peng, Mengchen Liu, Bin Xiao, Jianlong Fu, and Lu
  Yuan.
\newblock Tinyvit: Fast pretraining distillation for small vision transformers.
\newblock {\em arXiv preprint arXiv:2207.10666}, 2022.

\bibitem{detectron2}
Yuxin Wu, Alexander Kirillov, Francisco Massa, Wan-Yen Lo, and Ross Girshick.
\newblock Detectron2.
\newblock \url{https://github.com/facebookresearch/detectron2}, 2019.

\bibitem{resim}
Tete Xiao, Colorado~J Reed, Xiaolong Wang, Kurt Keutzer, and Trevor Darrell.
\newblock Region similarity representation learning.
\newblock In {\em Proceedings of the IEEE/CVF International Conference on
  Computer Vision}, pages 10539--10548, 2021.

\bibitem{orl}
Jiahao Xie, Xiaohang Zhan, Ziwei Liu, Yew~Soon Ong, and Chen~Change Loy.
\newblock Unsupervised object-level representation learning from scene images.
\newblock {\em Advances in Neural Information Processing Systems},
  34:28864--28876, 2021.

\bibitem{pixpro}
Zhenda Xie, Yutong Lin, Zheng Zhang, Yue Cao, Stephen Lin, and Han Hu.
\newblock Propagate yourself: Exploring pixel-level consistency for
  unsupervised visual representation learning.
\newblock In {\em Proceedings of the IEEE/CVF Conference on Computer Vision and
  Pattern Recognition}, pages 16684--16693, 2021.

\bibitem{simmim}
Zhenda Xie, Zheng Zhang, Yue Cao, Yutong Lin, Jianmin Bao, Zhuliang Yao, Qi
  Dai, and Han Hu.
\newblock Simmim: A simple framework for masked image modeling.
\newblock In {\em Proceedings of the IEEE/CVF Conference on Computer Vision and
  Pattern Recognition}, pages 9653--9663, 2022.

\bibitem{bingo}
Haohang Xu, Jiemin Fang, Xiaopeng Zhang, Lingxi Xie, Xinggang Wang, Wenrui Dai,
  Hongkai Xiong, and Qi Tian.
\newblock Bag of instances aggregation boosts self-supervised distillation.
\newblock In {\em International Conference on Learning Representations}, 2021.

\bibitem{cluster_fit}
Xueting Yan, Ishan Misra, Abhinav Gupta, Deepti Ghadiyaram, and Dhruv Mahajan.
\newblock Clusterfit: Improving generalization of visual representations.
\newblock In {\em Proceedings of the IEEE/CVF Conference on Computer Vision and
  Pattern Recognition}, pages 6509--6518, 2020.

\bibitem{instance_loc}
Ceyuan Yang, Zhirong Wu, Bolei Zhou, and Stephen Lin.
\newblock Instance localization for self-supervised detection pretraining.
\newblock In {\em Proceedings of the IEEE/CVF Conference on Computer Vision and
  Pattern Recognition}, pages 3987--3996, 2021.

\bibitem{lars}
Yang You, Igor Gitman, and Boris Ginsburg.
\newblock Large batch training of convolutional networks.
\newblock {\em arXiv preprint arXiv:1708.03888}, 2017.

\bibitem{attention_transfer}
Sergey Zagoruyko and Nikos Komodakis.
\newblock Paying more attention to attention: Improving the performance of
  convolutional neural networks via attention transfer.
\newblock {\em arXiv preprint arXiv:1612.03928}, 2016.

\bibitem{barlow_twins}
Jure Zbontar, Li Jing, Ishan Misra, Yann LeCun, and St{\'e}phane Deny.
\newblock Barlow twins: Self-supervised learning via redundancy reduction.
\newblock In {\em International Conference on Machine Learning}, pages
  12310--12320. PMLR, 2021.

\bibitem{rekd}
Kai Zheng, Yuanjiang Wang, and Ye Yuan.
\newblock Boosting contrastive learning with relation knowledge distillation.
\newblock In {\em Proceedings of the AAAI Conference on Artificial
  Intelligence}, volume~36, pages 3508--3516, 2022.

\end{thebibliography}
}

\newpage
\appendix
\section{Qualitative Results}

\begin{figure}[h]

    \centering

    \includegraphics[width=0.95\linewidth]{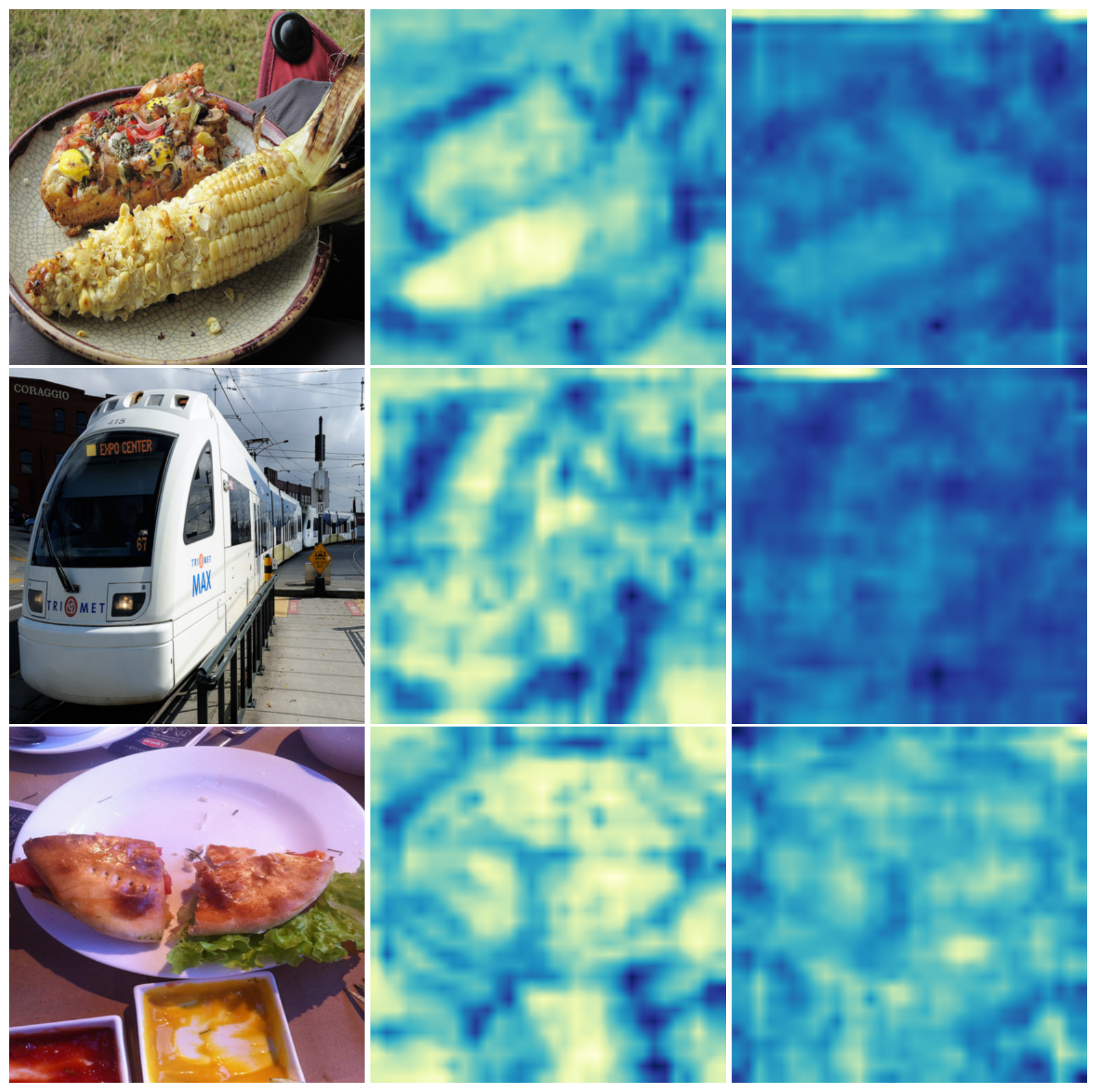}

    \caption{Visualization of feature maps. Images of the first column are from COCO \texttt{val2017}. The second and the third column depict the output feature maps of `layer4' of \mbox{ResNet-18} pre-trained by PCD and vectorized variant of PCD, respectively.}
    \label{fig1}
\end{figure}

In \cref{fig1}, we visualize the feature maps from `layer4' of \mbox{ResNet-18} pre-trained by PCD and vectorized variant of PCD, respectively. The feature maps of ResNet-18 pre-trained by PCD show clear outlines of input images. 

\section{Details of Fine-Tuning Experiments}

\subsection{Fine-Tuning on CityScapes}

The fine-tuning hyper-parameters are listed as follows:

\begin{center}
\tablestyle{8pt}{1.2}
\begin{tabular}{x{96}|x{68}}
Configuration & Value \\
\shline
optimizer & SGD \\
base learning rate & 0.01 \\
momentum & 0.9 \\
weight decay & 1e-4 \\
batch size & 16 \\
training steps & 90000 \\
learning rate schedule & WarmupMultiStepLR \\
warmup iters & 1000 \\
decay milestones & 63000, 81000 \\
shortest edge of resizing & $\left[512, 768, ..., 2048\right]$ \\
max input size & 4096 \\
random cropping & True \\
random flipping & True \\

\end{tabular}
\vspace{-.5em}
\label{tab:cityscapes_params} \vspace{-.5em}
\end{center} 
\hfill


\subsection{Fine-Tuning MobileNet v3 (Large)}

\mbox{MobileNet v3 (Large)} \cite{mobilenetv3} is not a stage-wise architecture like ResNet series \cite{resnet}. We have to manually define the ``stem'' and ``res'' stages in \mbox{MobileNet v3 (Large)} to fit in Detectron2. The rules for partitioning the modules of \mbox{MobileNet v3 (Large)} are: i) modules with the same stride belong to the same partition; ii) res4 must be of stride 16. We show the partitioning results:

\begin{center}

\tablestyle{8pt}{1.2}
\begin{tabular}{x{24}|x{24}|x{40}|x{24}}
index & modules & total stride & partitions \\
\shline
1 & conv & 2 & \multirow{2}{*}{stem} \\
2 & IBN & 2 \\
\hline
3 & IBN & 4 & \multirow{2}{*}{res2} \\
4 & IBN & 4 \\
\hline
5 & IBN & 8 & \multirow{3}{*}{res3} \\
6 & IBN & 8 \\
7 & IBN & 8 \\
\hline
8 & IBN & 16 & \multirow{6}{*}{res4} \\
9 & IBN & 16 \\
10 & IBN & 16 \\
11 & IBN & 16 \\
12 & IBN & 16 \\
13 & IBN & 16 \\
\hline
14 & IBN & 32 & \multirow{4}{*}{res5} \\
15 & IBN & 32 \\
16 & IBN & 32 \\
17 & conv & 32 \\

\end{tabular}
\vspace{-.5em}
\vspace{-.5em}
\end{center}
\hfill

\subsection{Distilling from Different Teachers}

In Sec.4.3, we try using other models as teachers. Most of checkpoints are from the official repository, except for BYOL \cite{byol}. The \mbox{ResNet-50} \cite{resnet} pre-trained by BYOL are from the implementation of \cite{revisiting_airl}. We list out the URLs for downloading these models:

\begin{itemize}
    \item \mbox{MoCo v3 (ResNet-50)} \cite{mocov3}: \url{https://dl.fbaipublicfiles.com/moco-v3/r-50-1000ep/r-50-1000ep.pth.tar}
    \item \mbox{MoCo v3 (ViT-Base)} \cite{mocov3}: \url{https://dl.fbaipublicfiles.com/moco-v3/vit-b-300ep/vit-b-300ep.pth.tar}
    \item SwAV \cite{swav}: \url{https://dl.fbaipublicfiles.com/deepcluster/swav_800ep_pretrain.pth.tar}
    \item BYOL \cite{byol}: \url{https://drive.google.com/file/d/1-5-O49vsro9YW9WTokSc8CoSrmjKfieB/view?usp=sharing}
    \item \mbox{Barlow Twins} \cite{barlow_twins}: \url{https://dl.fbaipublicfiles.com/barlowtwins/ljng/checkpoint.pth}
\end{itemize}

\mbox{Barlow Twins} has a projection head with an output dimension of $8192$. We have to use the variant of PCD with asymmetric loss to distill knowledge from \mbox{Barlow Twins} due to limited memory.

\end{document}